\documentclass[10pt,twocolumn,letterpaper]{article}

\usepackage{cvpr}              

\usepackage[dvipsnames]{xcolor}

\definecolor{cvprblue}{rgb}{0.21,0.49,0.74}
\usepackage[pagebackref,breaklinks,colorlinks,citecolor=cvprblue]{hyperref}
\usepackage{algorithm}
\usepackage{algorithmic}
\usepackage{csquotes}
\usepackage{multirow}
\usepackage{bm}        
\definecolor{coeff_a}{RGB}{245, 194, 66}
\definecolor{coeff_b}{RGB}{222, 131, 68}
\definecolor{coeff_c}{RGB}{79, 113, 190}

\def\paperID{6676} 
\def\confName{CVPR}
\def\confYear{2024}

\title{
\raisebox{-0.12cm}{\includegraphics[scale=0.18 ]{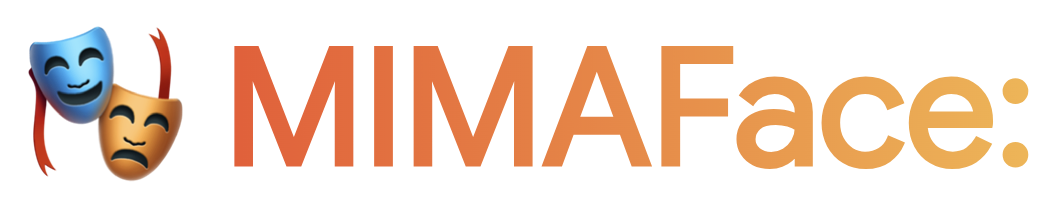}} Face Animation via Motion-Identity Modulated \\ Appearance Feature Learning}

\newcommand{\corresponding}{\textsuperscript{†}}

\author{
    Yue Han\textsuperscript{1}, 
    Junwei Zhu\textsuperscript{2}, 
    Yuxiang Feng\textsuperscript{1}, 
    Xiaozhong Ji\textsuperscript{2}, 
    Keke He\textsuperscript{2}, \\
    Xiangtai Li\textsuperscript{3}, 
    Zhucun Xue\textsuperscript{1}, 
    Yong Liu\corresponding\textsuperscript{1} \\
    \textsuperscript{1}Zhejiang University \quad 
    \textsuperscript{2}Tencent Youtu Lab \quad 
    \textsuperscript{3}Nanyang Technological University \\
    \{12432015, fengyx, 12432038\}@zju.edu.cn, yongliu@iipc.zju.edu.cn \\
    \url{https://mimaface2024.github.io/mimaface.github.io}
} 

\begin{document}
\maketitle
\renewcommand\thefootnote{\fnsymbol{footnote}}
\footnotetext[1]{\corresponding denotes corresponding author.}

\def\mA{\mathcal{A}}
\def\mB{\mathcal{B}}
\def\mC{\mathcal{C}}
\def\mD{\mathcal{D}}
\def\mE{\mathcal{E}}
\def\mF{\mathcal{F}}
\def\mG{\mathcal{G}}
\def\mH{\mathcal{H}}
\def\mI{\mathcal{I}}
\def\mJ{\mathcal{J}}
\def\mK{\mathcal{K}}
\def\mL{\mathcal{L}}
\def\mM{\mathcal{M}}
\def\mN{\mathcal{N}}
\def\mO{\mathcal{O}}
\def\mP{\mathcal{P}}
\def\mQ{\mathcal{Q}}
\def\mR{\mathcal{R}}
\def\mS{\mathcal{S}}
\def\mT{\mathcal{T}}
\def\mU{\mathcal{U}}
\def\mV{\mathcal{V}}
\def\mW{\mathcal{W}}
\def\mX{\mathcal{X}}
\def\mY{\mathcal{Y}}
\def\mZ{\mathcal{Z}} 

\def\bbN{\mathbb{N}} 
\def\bbR{\mathbb{R}} 
\def\bbP{\mathbb{P}} 
\def\bbQ{\mathbb{Q}} 
\def\bbE{\mathbb{E}}

\def\1n{\mathbf{1}_n}
\def\0{\mathbf{0}}
\def\1{\mathbf{1}}

\def\A{{\bf A}}
\def\B{{\bf B}}
\def\C{{\bf C}}
\def\D{{\bf D}}
\def\E{{\bf E}}
\def\F{{\bf F}}
\def\G{{\bf G}}
\def\H{{\bf H}}
\def\I{{\bf I}}
\def\J{{\bf J}}
\def\K{{\bf K}}
\def\L{{\bf L}}
\def\M{{\bf M}}
\def\N{{\bf N}}
\def\O{{\bf O}}
\def\P{{\bf P}}
\def\Q{{\bf Q}}
\def\R{{\bf R}}
\def\S{{\bf S}}
\def\T{{\bf T}}
\def\U{{\bf U}}
\def\V{{\bf V}}
\def\W{{\bf W}}
\def\X{{\bf X}}
\def\Y{{\bf Y}}
\def\Z{{\bf Z}}

\def\a{{\bf a}}
\def\b{{\bf b}}
\def\c{{\bf c}}
\def\d{{\bf d}}
\def\e{{\bf e}}
\def\f{{\bf f}}
\def\g{{\bf g}}
\def\h{{\bf h}}
\def\i{{\bf i}}
\def\j{{\bf j}}
\def\k{{\bf k}}
\def\l{{\bf l}}
\def\m{{\bf m}}
\def\n{{\bf n}}
\def\o{{\bf o}}
\def\p{{\bf p}}
\def\q{{\bf q}}
\def\r{{\bf r}}
\def\s{{\bf s}}
\def\t{{\bf t}}
\def\u{{\bf u}}
\def\v{{\bf v}}
\def\w{{\bf w}}
\def\x{{\bf x}}
\def\y{{\bf y}}
\def\z{{\bf z}}

\def\balpha{\mbox{\boldmath{$\alpha$}}}
\def\bbeta{\mbox{\boldmath{$\beta$}}}
\def\bdelta{\mbox{\boldmath{$\delta$}}}
\def\bgamma{\mbox{\boldmath{$\gamma$}}}
\def\blambda{\mbox{\boldmath{$\lambda$}}}
\def\bsigma{\mbox{\boldmath{$\sigma$}}}
\def\btheta{\mbox{\boldmath{$\theta$}}}
\def\bomega{\mbox{\boldmath{$\omega$}}}
\def\bxi{\mbox{\boldmath{$\xi$}}}
\def\bnu{\mbox{\boldmath{$\nu$}}}                                  
\def\bphi{\mbox{\boldmath{$\phi$}}}
\def\bmu{\mbox{\boldmath{$\mu$}}}

\def\bDelta{\mbox{\boldmath{$\Delta$}}}
\def\bOmega{\mbox{\boldmath{$\Omega$}}}
\def\bPhi{\mbox{\boldmath{$\Phi$}}}
\def\bLambda{\mbox{\boldmath{$\Lambda$}}}
\def\bSigma{\mbox{\boldmath{$\Sigma$}}}
\def\bGamma{\mbox{\boldmath{$\Gamma$}}}
                                  
\newcommand{\myprob}[1]{\mathop{\mathbb{P}}_{#1}}

\newcommand{\myexp}[1]{\mathop{\mathbb{E}}_{#1}}

\newcommand{\mydelta}[1]{1_{#1}}

\newcommand{\myminimum}[1]{\mathop{\textrm{minimum}}_{#1}}
\newcommand{\mymaximum}[1]{\mathop{\textrm{maximum}}_{#1}}    
\newcommand{\mymin}[1]{\mathop{\textrm{minimize}}_{#1}}
\newcommand{\mymax}[1]{\mathop{\textrm{maximize}}_{#1}}
\newcommand{\mymins}[1]{\mathop{\textrm{min.}}_{#1}}
\newcommand{\mymaxs}[1]{\mathop{\textrm{max.}}_{#1}}  
\newcommand{\myargmin}[1]{\mathop{\textrm{argmin}}_{#1}} 
\newcommand{\myargmax}[1]{\mathop{\textrm{argmax}}_{#1}} 
\newcommand{\myst}{\textrm{s.t. }}

\newcommand{\denselist}{\itemsep -1pt}
\newcommand{\sparselist}{\itemsep 1pt}

\definecolor{pink}{rgb}{0.9,0.5,0.5}
\definecolor{purple}{rgb}{0.5, 0.4, 0.8}   
\definecolor{gray}{rgb}{0.3, 0.3, 0.3}
\definecolor{mygreen}{rgb}{0.2, 0.6, 0.2}

\newcommand{\cyan}[1]{\textcolor{cyan}{#1}}
\newcommand{\blue}[1]{\textcolor{blue}{#1}}
\newcommand{\magenta}[1]{\textcolor{magenta}{#1}}
\newcommand{\pink}[1]{\textcolor{pink}{#1}}
\newcommand{\green}[1]{\textcolor{green}{#1}} 
\newcommand{\gray}[1]{\textcolor{gray}{#1}}    
\newcommand{\mygreen}[1]{\textcolor{mygreen}{#1}}    
\newcommand{\purple}[1]{\textcolor{purple}{#1}}       

\definecolor{greena}{rgb}{0.4, 0.5, 0.1}
\newcommand{\greena}[1]{\textcolor{greena}{#1}}

\definecolor{bluea}{rgb}{0, 0.4, 0.6}
\newcommand{\bluea}[1]{\textcolor{bluea}{#1}}
\definecolor{reda}{rgb}{0.6, 0.2, 0.1}
\newcommand{\reda}[1]{\textcolor{reda}{#1}}

\def\changemargin#1#2{\list{}{\rightmargin#2\leftmargin#1}\item[]}
\let\endchangemargin=\endlist
                                               
\newcommand{\cm}[1]{}

\newcommand{\mhoai}[1]{{\color{magenta}\textbf{[MH: #1]}}}
\newcommand{\ruoyux}[1]{{\color{purple}\textbf{[RX: #1]}}}

\newcommand{\mtodo}[1]{{\color{red}$\blacksquare$\textbf{[TODO: #1]}}}
\newcommand{\myheading}[1]{\vspace{1ex}\noindent \textbf{#1}}
\newcommand{\htimesw}[2]{\mbox{$#1$$\times$$#2$}}

\newcommand{\young}[1]{{\color{blue}$\blacksquare$\textbf{Alternative}: #1}}


\newif\ifshowsolution
\showsolutiontrue

\ifshowsolution  
\newcommand{\Comment}[1]{\paragraph{\bf $\bigstar $ COMMENT:} {\sf #1} \bigskip}
\newcommand{\Solution}[2]{\paragraph{\bf $\bigstar $ SOLUTION:} {\sf #2} }
\newcommand{\Mistake}[2]{\paragraph{\bf $\blacksquare$ COMMON MISTAKE #1:} {\sf #2} \bigskip}
\else
\newcommand{\Solution}[2]{\vspace{#1}}
\fi

\newcommand{\truefalse}{
\begin{enumerate}
	\item True
	\item False
\end{enumerate}
}

\newcommand{\yesno}{
\begin{enumerate}
	\item Yes
	\item No
\end{enumerate}
}

\newcommand{\Sref}[1]{Sec.~\ref{#1}}
\newcommand{\Eref}[1]{Eq.~(\ref{#1})}
\newcommand{\Fref}[1]{Fig.~\ref{#1}}
\newcommand{\Tref}[1]{Table~\ref{#1}}
\begin{abstract}
Current diffusion-based face animation methods generally adopt a ReferenceNet (a copy of U-Net) and a large amount of curated self-acquired data to learn appearance features, as robust appearance features are vital for ensuring temporal stability. 
However, when trained on public datasets, the results often exhibit a noticeable performance gap in image quality and temporal consistency. 
To address this issue,  we meticulously examine the essential appearance features in the facial animation tasks, which include motion-agnostic (e.g., clothing, background) and motion-related (e.g., facial details) texture components, along with high-level discriminative identity features.
Drawing from this analysis, we introduce a Motion-Identity Modulated Appearance Learning Module (MIA) that modulates CLIP features at both motion and identity levels. 
Additionally, to tackle the semantic/ color discontinuities between clips, we design an Inter-clip Affinity Learning Module (ICA) to model temporal relationships across clips.
Our method achieves precise facial motion control (i.e., expressions and gaze), faithful identity preservation, and generates animation videos that maintain both intra/inter-clip temporal consistency. 
Moreover, it easily adapts to various modalities of driving sources. Extensive experiments demonstrate the superiority of our method.
%
\end{abstract}

\section{Introduction}
\label{sec:introduction}


Face Animation aims to generate a realistic talking head by animating a source face using the motion information of another person, \ie, pose, expression, and gaze~\cite{pei2024deepfake}.
It has diverse applications in virtual character creation for game production and video editing. 
Previous GAN-based methods ~\cite{monkeynet, freenet, xu2022designing, nirkin2019fsgan, fomm,pirenderer,yang2022face2face,styleheat,bounareli2023hyperreenact,zhang2023metaportrait,dagan,dam,mcnet} most delivers results at resolution $256^2$ and only support median pose variation, \ie, less than 30 degrees), due to the lack of high-resolution and large pose dataset. 
Recent attempts leverage the powerful generation capability of pre-trained latent diffusion models to address these challenges. 
However, the high variance of noise in diffusion presents a new challenge in generating smooth videos.

\begin{figure}[tp]
    \centering
    \includegraphics[width=1.0\linewidth]{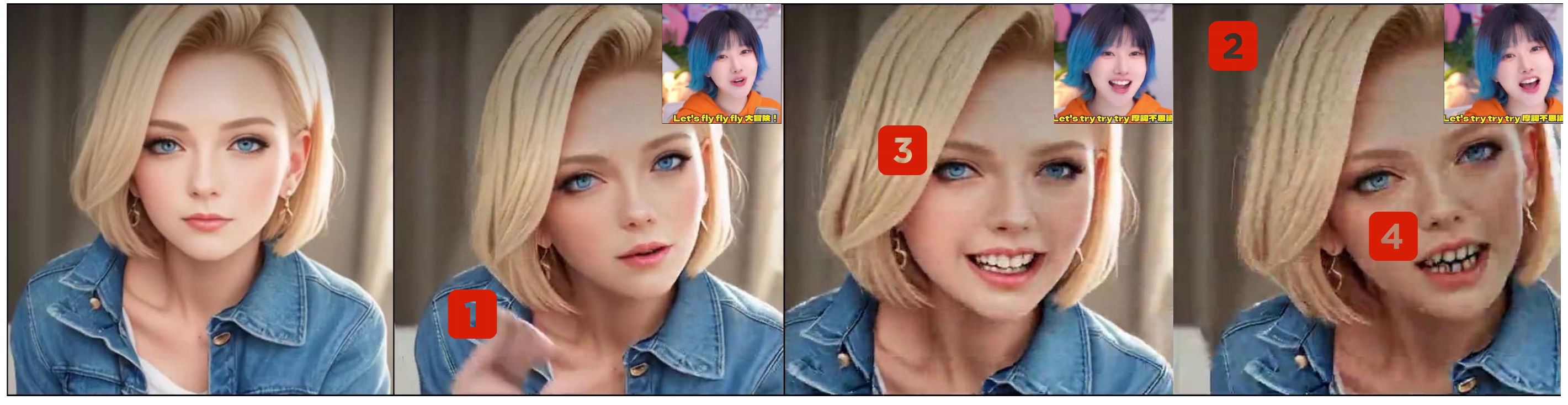}
    \caption{
    Typical failure cases for current diffusion-based face animation methods: \textbf{(1)/(2)} semantic/ color discontinuity across clips, \textbf{(3)} stiff expression, \textbf{(4)} quality degradation}   
    \label{fig:motivation}
    \vspace{-6mm}
\end{figure}

The human-body animation methods~\cite{animateanyone,magicanimate} identify that CLIP~\cite{clip} fails to provide adequate appearance features, resulting in video flickering. 
To address this issue, ReferenceNet (a UNet copy) is proposed to supply multi-scale similar appearance features.
Subsequent works in face animation adopt similar frameworks, integrating ReferenceNet with motion modules proposed in AnimateDiff~\cite{animatediff} to ensure temporal stability and achieve remarkably favorable results. 
However, the quality of the results within this paradigm ~\cite{animateanyone, magicanimate,emo,magicdance,aniportrait} heavily depends on the quality and volume of the training data. When trained on publicly available datasets~\cite{magicdance,aniportrait}, the generated results often exhibit a noticeable gap in quality, as shown in \cref{fig:motivation} (from released examples in AniPortrait). 
A question arises: \textit{what appearance features are essential for producing a temporally stable, high-quality face animation video?} 
We argue that the necessary appearance features include motion-independent texture features (\ie, clothing, background), motion-related texture features (\ie, facial details), and high-level discriminative features (\ie, identity). 
Based on this analysis, we propose the Motion-Identity Modulated Appearance Learning Module (MIA), which modulates CLIP features at both motion and identity levels. 
For motion modulation, we use 3DMM coefficients to modulate appearance features via cross-attention. This facilitates the generation of subtle facial textures, \eg, wrinkles and muscle contractions, resulting from expressions. For identity modulation, we introduce an identity contrastive loss to compensate for high-level discriminative features lost. This is because we find that the optimization objective, i.e., the denoise loss, encourages CLIP to focus more on the learning of low-level generative features while neglecting the learning of discriminative features. 
This necessitates that the model learns to preserve identity through training on data that encompasses a wider range of identities.
The joint modulation of motion and identity makes appearance features learning easier, allowing us to achieve high-fidelity and temporally stable video results even when trained solely on public datasets.

Although adequate appearance features ensure intra-clip temporal consistency with temporal attention, the lack of modeling inter-clip relationships causes semantic/ color discontinuities between clips. 
Current solutions include fixing initial noise, frame interpolation, or temporal co-denoising, but noticeable discontinuities still exist. 
To address this issue, we introduce the Inter-clip Affinity Learning Module (ICA), which conditions preceding frames with augmentation strategies to bridge the gap between ground truth frames during training and generated frames during inference. 

Our contributions can be summarized as follows:
\begin{itemize}
  \item 
    We identify potential issues in current diffusion-based face animation methods and carefully examine the essential appearance features. Based on the analysis, we introduce a Motion-Identity Modulated Appearance Learning Module (MIA) that modulates CLIP features at both motion and identity levels, making the appearance features learning more effective.
  \item We introduce the Inter-clip Affinity Learning Module (ICA), which models temporal relationships between clips to address the issue of semantic/color discontinuities.
  \item Our method achieves precise facial motion control (i.e. expressions and gaze), faithful identity preservation, and generates animation videos that maintain both intra/inter-clip temporal consistency. Moreover, it easily adapts to various modalities of driving sources. Extensive experiments demonstrate the superiority of our method.
\end{itemize}

\begin{figure}[tp]
    \centering
    \includegraphics[width=1.0\linewidth]{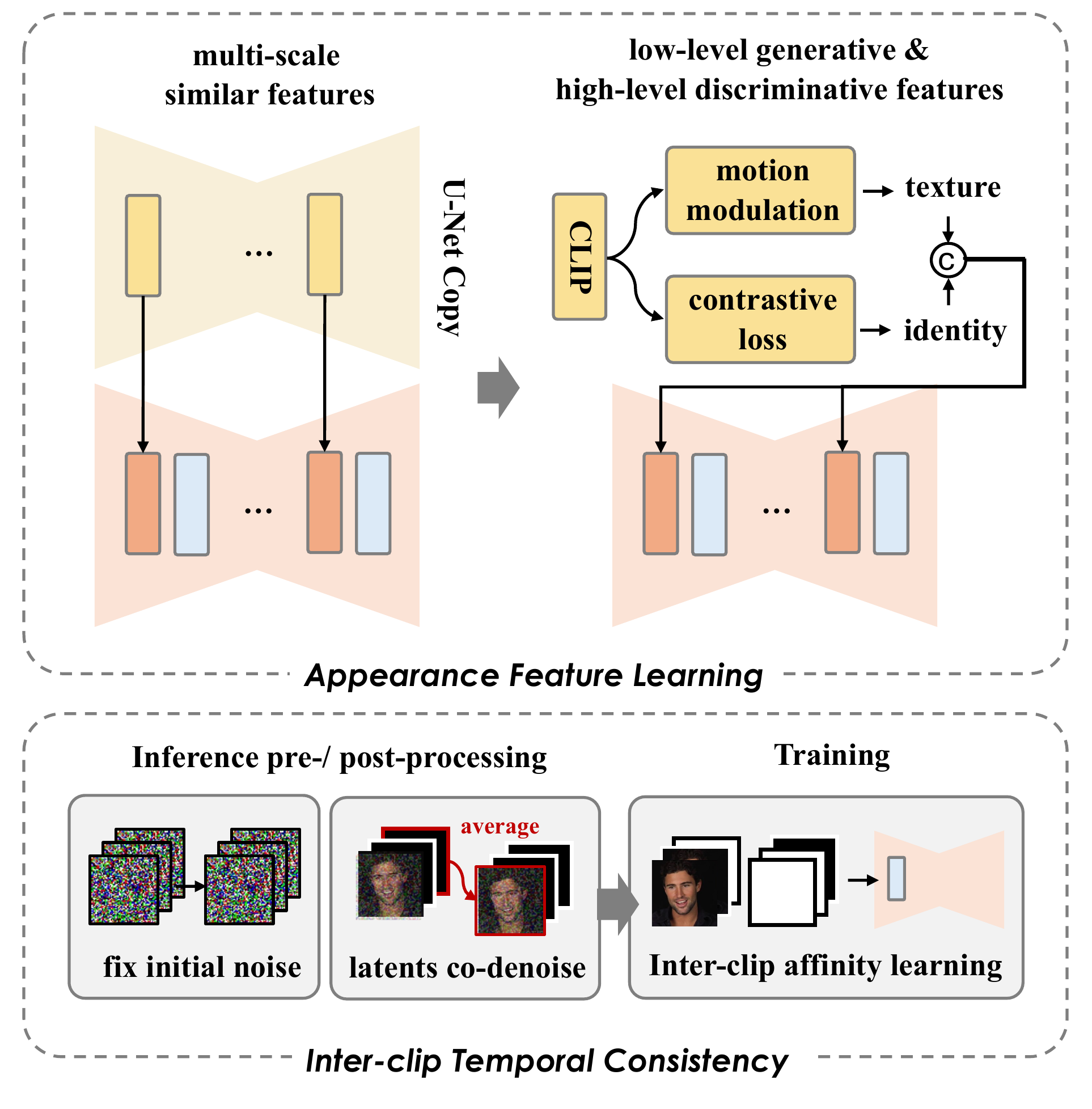}
    \caption{
        We compare our method to previous diffusion-based face animation methods in terms of appearance feature learning and inter-clip temporal consistency. 
        }   
    \label{fig:intro}
\end{figure}

\section{Related Works}
\label{sec:related_works}

\noindent
\subsection{GAN-based Face Animation}
Methods~\cite{monkeynet, freenet, xu2022designing, nirkin2019fsgan, agarwal2023audio, xu2022region, yang2022face2face,bounareli2023hyperreenact,zhang2023metaportrait,gao2023high,bounareli2024one} can be broadly divided into warping-based and 3DMM-based methods. 
\textit{Warping-based methods} ~\cite{monkeynet,tpsm,fomm,dagan}typically extract landmarks or region pairs to estimate flow fields and perform warping on the source appearance feature maps to transfer motions.
%
%
%
Limited by the accuracy of the predicted flow field, these methods tend to produce blurry and distorted results when dealing with large motion variations.
%

\textit{3DMM-based methods} use facial reconstruction coefficients or the render image from 3DMM as motion intermediate representation. 
Due to the inherent decoupling properties of coefficients, 3DMM-based methods, \eg, PIRenderer~\cite{pirenderer}, can freely control expressions and poses.
Although 3DMM provides accurate structural references for facial regions, it lacks references for hair, teeth, and eye movement.  
Additionally, its coarse facial textures result in suboptimal generated outcomes. 
StyleHEAT~\cite{styleheat} tackles this challenge by leveraging the powerful generation capabilities of StyleGAN2 to produce detailed textures and high-resolution frontal portraits. 
However, its generalization and large pose generation abilities are limited due to the constraint of the training dataset.

\noindent
\subsection{Diffusion-based Face Animation}
To improve sample quality and generalization capability, diffusion models have gained popularity~\cite{fadm,peng2023portraitbooth,animateanyone,magicanimate,han2023generalist}.  
FADM~\cite{fadm} combines the previous reenactment models with diffusion refinements, but the base model limits the driving accuracy. 
The human-body animation methods~\cite{animateanyone,magicanimate} identify that CLIP fails to provide adequate appearance features, resulting in video flickering. 
To address this issue, ReferenceNet (a UNet copy) is proposed to supply multi-scale similar appearance features.
Subsequent works in face animation adopt similar frameworks, integrating ReferenceNet with motion modules proposed in AnimateDiff~\cite{animatediff} to ensure temporal stability and achieve remarkably favorable results. 
However, the quality of the results within this paradigm ~\cite{animateanyone, magicanimate,emo,magicdance,aniportrait} heavily depends on the quality and volume of the training data. When trained on publicly available datasets~\cite{magicdance,aniportrait}, the generated results often exhibit a noticeable gap in quality.
 To address this issue, in this paper, we explore the effective way to learn robust appearance features.

%
%

\noindent
\subsection{Audio-Driven Face Animation} 
%
Previous approaches ~\cite{wav2lip,makelttalk, zhang2020apb2face, zhang2021real, fan2022faceformer, xing2023codetalker, xu2023multimodal, garcia2023cognitive, shen2022learning, huang2023parametric, daetalker, tan2024flowvqtalker, ye2024realdportrait, wang2024styletalkplus} focus on learning models specific to individual speakers. 
%
%
Sadtalker~\cite{sadtalker} uses 3DMM as an intermediate representation for subject-agnostic reenactment. 
Via learning the 3D motion coefficients of the 3DMM model from audio, Sadtalker showcases robust generalization capabilities. 
%
However, it still struggles to accommodate significant motion variations and produces blurry results.
Recently, several works~\cite{emo, aniportrait} have followed the 'ReferenceNet with motion module' paradigm and achieved significant advancements.
~\cite{emo} bypasses the need for intermediate 3D models or facial landmarks and achieve astonishing results. However, it heavily relies on the dataset, making it difficult to reproduce.
~\cite{aniportrait} learns to map the audio to a 3D facial mesh and head pose, then projects these two elements into 2D keypoints as the intermediate motion representation. However, it also exhibits typical failure cases similar to methods using the same paradigm.
In this paper, our primary goal is to explore how to effectively learn robust appearance features, not to focus on the learning of motion intermediate representations. Following the approach of ~\cite{styleheat, aniportrait}, which can be easily adapted to multi-modal inputs, we support audio-driven face animation by mapping audio to the 3DMM coefficient space using pre-trained audio-to-3DMM-coefficients encoders, \,  e.g., ~\cite{sadtalker}.

\noindent
\subsection{Temporal Consistency in Face Animation.}
Current methods~\cite{magicanimate, magicdance, emo, aniportrait, wu2024motionbooth, wu2024lgvi} tend to focus more on modeling the temporal stability within clips while neglecting the preservation of temporal consistency between clips. Recent image animation~\cite{pia} methods (typically generating only one clip) demonstrate that temporal attention is sufficient for ensuring intra-clip temporal stability. Therefore, we argue the temporal stability within clips is mainly affected by the lack of adequate appearance features. Current solutions for inter-clip temporal stability include fixing initial noise, frame interpolation, or temporal co-denoising~\cite{magicanimate}, but noticeable discontinuities still persist. Unlike these pre/post-processing approaches during inference, we explicitly model the temporal relationships between clips during training.


%

\section{Methods}
\label{sec:methods}

\begin{figure*}[t]
    \centering
    \includegraphics[width=1.0\linewidth]{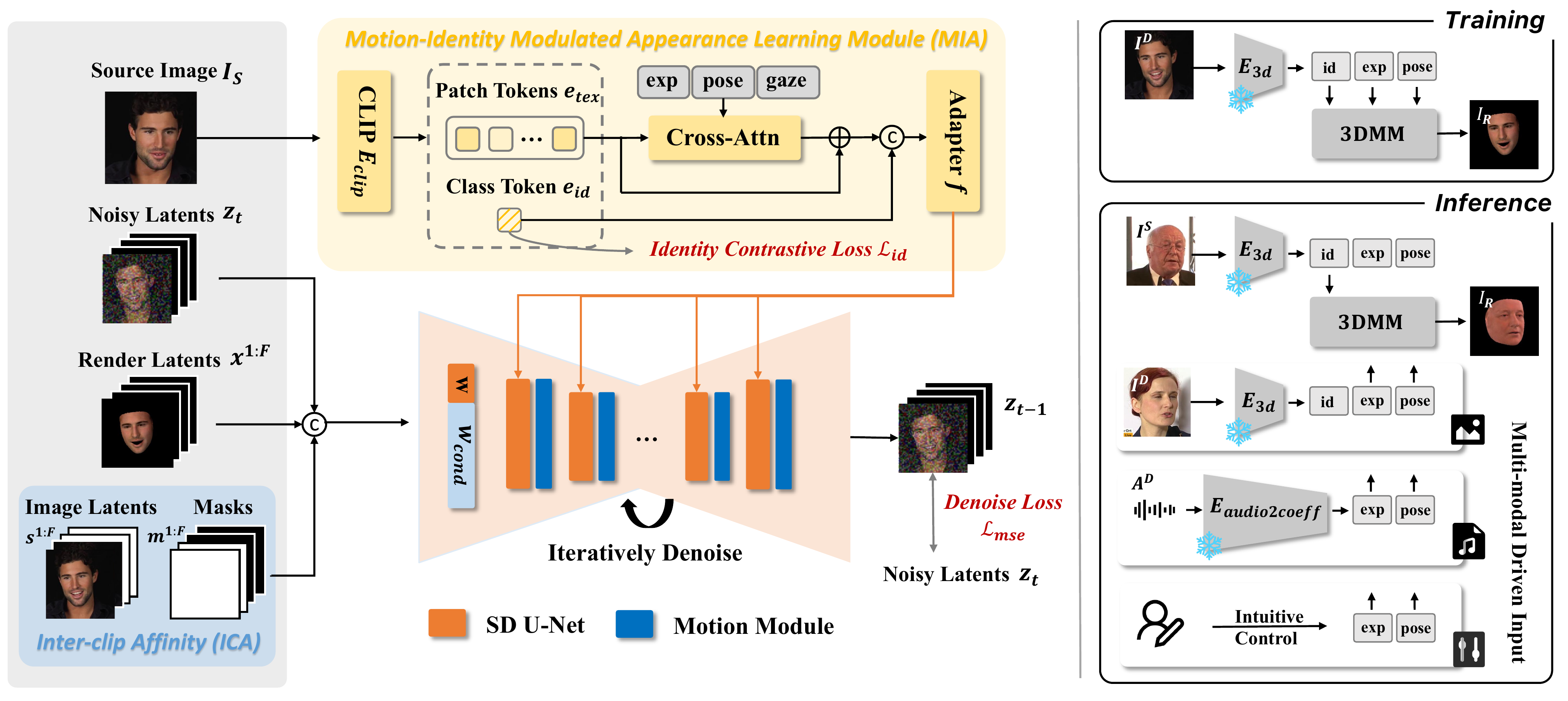}
    \caption{\textbf{Pipeline of the proposed MIMAFace}, which consists of: \textbf{\textit{1)}} \textcolor{coeff_a}{\textbf{Motion-Identity Modulated Appearance Learning Module (MIA)}} and \textbf{\textit{2)}} \textcolor{coeff_c}{\textbf{Inter-clip Affinity Learning Module (ICA)}}. MIA modulates the appearance features at both motion and identity levels. The source image $\boldsymbol{I}_{S}$ is passed to CLIP $\boldsymbol{E}_{clip}$ to obtain patch tokens $\bm{e}_{tex}$ and a class token $\bm{e}_{id}$, which capture the texture and identity, respectively. $\bm{e}_{tex}$ are then modulated with motion coefficients $\boldsymbol{\rho}, \boldsymbol{\beta}, \boldsymbol{g}$ via cross attention. $\bm{e}_{id}$ is used to calculate the identity contrastive loss $\mathcal{L}_{id}$. The modulated  $\bm{e}_{tex}$ are concatenated with $\bm{e}_{id}$ to form the conditioning appearance features.
    ICA ensures inter-clip temporal consistency by conditioning image latent $\boldsymbol{s}^{1: F}$ (of ground truth during training and denoised ones during inference) and indicating masks $\boldsymbol{m}^{1: F}$ with the added condition module $\boldsymbol{W}_{cond}$ .
    Additionally, we employ 3DMM coefficients $\boldsymbol{\rho}, \boldsymbol{\beta}$ and rendered images $\boldsymbol{I}_{R}$ as intermediate representations for motion. The 3DMM coefficients can adapt to various modalities of driving inputs, \ie, images, audio, and manual modifications.
    }
    \label{fig:pipeline}
\end{figure*}


Face Animation aims to create a lifelike talking head video by animating a source face $\boldsymbol{I}_{S}$ using motion information from another person. 
This motion information may include pose, expression, and gaze, which can be obtained from either a video sequence $\boldsymbol{I}_{D}^{1:N}$ or an audio sequence $\boldsymbol{A}_D^{1:N}$. 
The entire sequence has a length of $\boldsymbol{N}$ and is divided into several clips with the length of $\boldsymbol{F}$. 

The current challenge in generating high-quality and smooth animation videos using diffusion-based methods can be summarized as the need for robust appearance features and ensuring temporal consistency within and between clips:
\textbf{\textit{1)}} 
Recent methods commonly use ReferenceNet for extracting appearance features, but this approach relies on learning from large amounts of curated data. 
We believe this is due to the lack of guidance on effective optimization directions. Thus, we propose MIA that modulates appearance features at both motion and identity levels to address this issue. The motion modulation encourages the model to learn facial texture details related to motion, while the identity modulation focuses on high-level discriminative features beyond low-level texture information guided by denoise loss.
\textbf{\textit{2)}}
To ensure temporal consistency, the motion module is shown to maintain intra-clip temporal consistency effectively. 
However, the challenge of ensuring inter-clip consistency remains largely unexplored. 
Existing methods~\cite{magicanimate, aniportrait} like frame interpolation, fixed initial noise, and temporal co-denoising have proven insufficient in resolving flickering. 
In this work, we propose ICA to address this issue by modeling relationships with preceding frames during training.
The pipeline of the proposed MIMAFace is illustrated in 
\cref{fig:pipeline}

\subsection{Motion Intermediate Representation}

We utilize the motion coefficients of 3DMM, \ie, pose $\boldsymbol{\rho}$, expression $\boldsymbol{\beta}$, and the rendered image $\boldsymbol{I}_{R}$, as a unified motion intermediate representation (for details on 3DMM, please refer to the supplementary materials). 
The rendered image $\boldsymbol{I}_{R}$ is conditioned by concatenating with the noisy latent $\bm{z}_{t}$. 
The coefficients are used to modulate the appearance features to supplement more motion-related facial details.
This approach offers several advantages:
\textbf{\textit{1)}} Highly disentangled coefficient space allows for convenient integration of multi-modal driving sources, mapping images/ audio to 3D coefficients with pre-trained models;
\textbf{\textit{2)}} Continuous coefficient space facilitates smooth video generation, in contrast to the spatial jittering characteristics of landmarks;
\textbf{\textit{3)}} The rendered image $\boldsymbol{I}_{R}$, similar to landmarks, provides coarse spatial guidance, while the coefficient encodes more information, compensating for the missing details about subtle facial expressions.
Besides 3DMM, we employ an off-the-shelf gaze detector to extract gaze embedding $\boldsymbol{g}$.


\subsection{Motion-Identity Modulated Appearance Learning Module (MIA)}
\label{sec:mia}

The robust appearance features should include motion-independent texture features (\ie, clothing, background), motion-related texture features (\ie, facial details), and high-level discriminative features (\ie, identity). 
However, under the guidance of the denoise loss, the appearance encoder tends to learn common textures more easily. This leads to two issues: 1) Although the appearance interacts with the motion intermediate representation through SD cross-attention, it is difficult for landmarks/render to express subtle facial movements. 2) The optimization target for generation enc, we introduce modulation at both the motion and identity levels to improve the effectiveness and robustness of appearance featuresslow convergence of the model. As a result, to improve the effectiveness and robustness of appearance features, we introduce modulation at both the motion and identity levels.
Specifically, we still choose CLIP as the appearance encoder to validate our hypothesis. The source image is passed through the CLIP vision encoder $\boldsymbol{E}_{clip}$, yielding patch tokens $\bm{e}_{tex}$ to capture low-level textures, and a class token $\bm{e}_{id}$ to represent high-level identity.

\noindent
\textbf{Motion Modulation}
To enable the model to generate more vivid and subtle facial movements, we introduce additional 3DMM motion parameters $\boldsymbol{\rho}$, $\boldsymbol{\beta}$ and gaze embedding $\boldsymbol{g}$ to modulate the texture tokens $\bm{e}_{tex}$ via simple cross-attention with a residual connection to prevent information loss:
\begin{equation}
\scalebox{0.8}{$
\begin{aligned}
 \boldsymbol{e}_{t e x}^{\prime}=\boldsymbol{e}_{t e x}+\operatorname{Cross-Attn}\left(\boldsymbol{e}_{t e x},[\boldsymbol{\rho} ; \boldsymbol{\beta} ; \boldsymbol{g}]\right), \\
 \operatorname{Cross-Attn}\left(\boldsymbol{e}_{t e x},[\boldsymbol{\rho} ; \boldsymbol{\beta} ; \boldsymbol{g}]\right)=\operatorname{softmax}\left(\frac{\boldsymbol{e}_{t e x} \cdot[\boldsymbol{\rho} ; \boldsymbol{\beta} ; \boldsymbol{g}]^T}{\sqrt{d_k}}\right)[\boldsymbol{\rho} ; \boldsymbol{\beta} ; \boldsymbol{g}],
\end{aligned}
$}
\end{equation}
where $d_k$ is the dimensionality of the key.
This process can be understood as adding more motion information to refine the coarse spatial structure determined by the render. It can also be seen as aligning the facial appearance features to the same motion as the render, making the cross-attention feature interaction within SD more effective.

\begin{figure}[tp]
    \centering
    \includegraphics[width=1.0\linewidth]{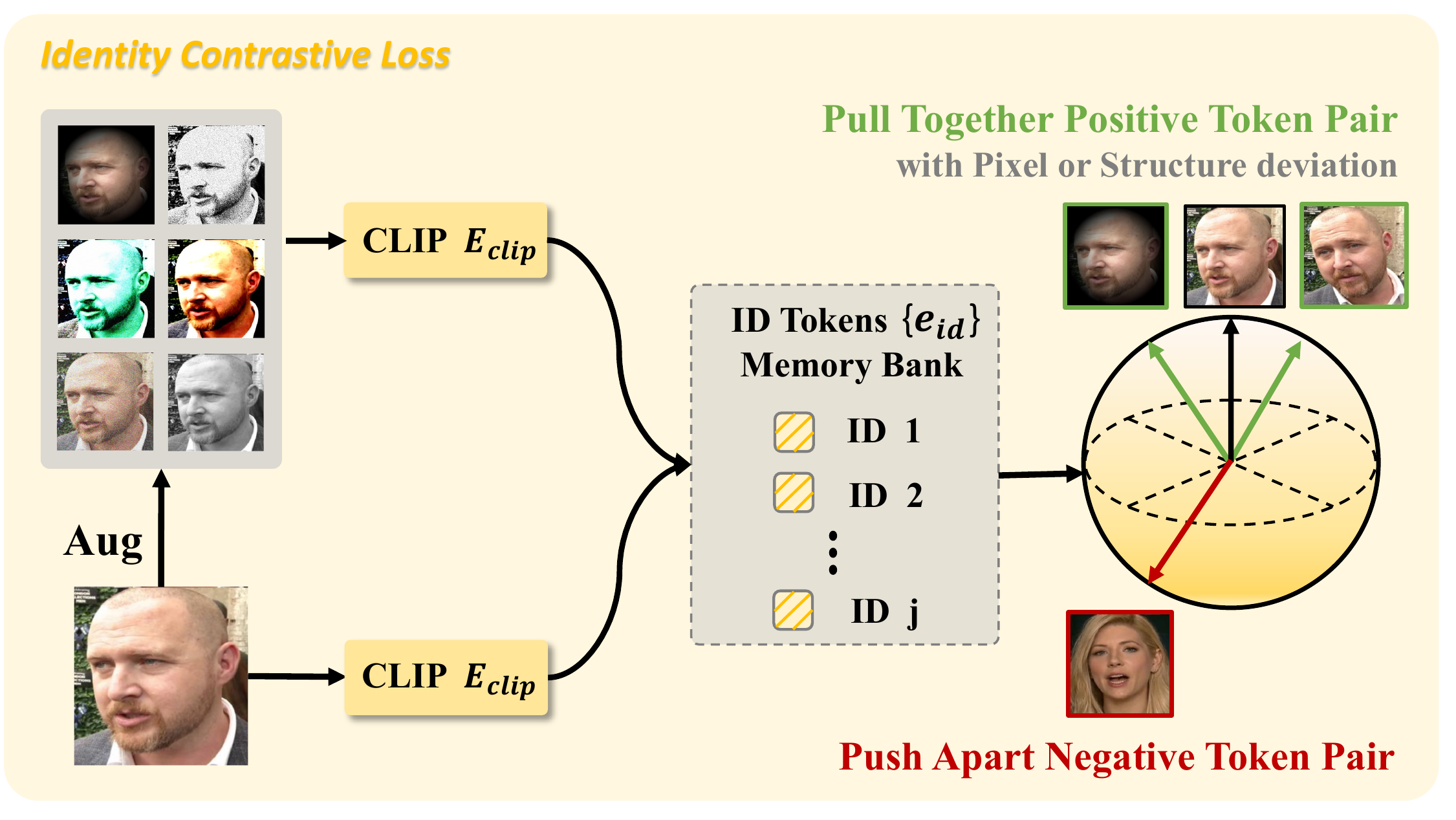}
    \vspace{-3mm}
    \caption{
    \textbf{Illustration of our Identity Contrastive Loss.}
    We apply photometric data augmentation on the source image and maintain an ID token memory bank to store ID tokens. 
    By pulling together positive token pairs with variances in pixels or structure and pushing apart the negative token pairs, the loss encourages the appearance encoder to capture high-level discriminative features.
    }
    \label{fig:idloss}
    \vspace{-1em}
\end{figure}

\noindent
\textbf{Identity Modulation}
To prevent the model from neglecting the learning of high-level features, we introduce a discriminative objective. For faces, identity is the most intuitive high-level feature. Therefore, we incorporate an \textit{Identity Contrastive Loss} $\mathcal{L}_{\text{id}}$ during training, as illustrated in \cref{fig:idloss}. 
Specifically, we augment the source image $\boldsymbol{I}_{S}$ through photometric transformation, altering the RGB channels by shifting pixel colors to new values. 
This includes techniques such as grayscaling, color jittering, various filtering methods (like edge enhancement, blurring, and sharpening), lighting perturbation, noise addition, vignetting, and contrast adjustment.
Subsequently, the source image $\boldsymbol{I}_{S}$ and the augmented image $\boldsymbol{I}_{aug}$ are each processed through $\boldsymbol{E}_{clip}$ to obtain a positive ID token pair. 
During the training process, we maintain an ID token memory bank $\mathcal{M}=\{\bm{e}_{id}\}$ to store ID tokens. 
Consequently, an ID token belonging to the same person but with a different structure will produce another type of positive ID token pair. 
Explicitly combining the same identity samples across pixel and structural variations enhances the model's generalization capability and robustness.
The Identity Contrastive Loss $\mathcal{L}_{id}$ is given by:
\begin{equation}
\resizebox{0.99\linewidth}{!}{
    $\mathcal{L}_{id} = -\log \left(\frac{\exp \left(\operatorname{sim}\left(z_i, z_i^{+}\right)\right)}{\exp \left(\operatorname{sim}\left(z_i, z_i^{+}\right)\right)+\sum_{j=1}^N \exp \left(\operatorname{sim}\left(z_i, z_j^{-}\right)\right)}\right)$,
}
\end{equation}
where $z_i$ represents the token of the identity token, $z_i^{+}$ and $z_j^{-}$ represent the token of the same and different identity, respectively.
$\operatorname{sim}\left(z_i, z_j\right)$ denotes the cosine similarity between two tokens. 
Finally, the modulated patch tokens $\bm{e'}_{tex}$ and the class token $\bm{e}_{id}$ are concatenated and go through an adapter $\bm{f}$ constructed by a 3-layer transformer to obtain the final appearance condition.

\subsection{Inter-clip Affinity learning Module (ICA)}
\label{sec:ica}
Although temporal consistency within video clips is learned and preserved through temporal attention layers, variations in lighting, color, and semantics may still occur between different clips. 
Previous methods ~\cite{magicanimate, aniportrait} commonly apply noise pre-processing or denoised latent post-processing during inference without explicitly learning temporal correspondence. 

Our goal is to enable the model to smoothly generate the next video clip by referring to the last few frames in the previous clip.
This problem can be reformulated as follows: When the model is conditioned on the preceding frames, we aim for it to reconstruct these frames, thereby maintaining temporal consistency with the preceding frames. 
This is because the temporal consistency of subsequent frames within the video clip can be ensured by temporal attention. 
During training, given a video clip of length $F$, the initial $k$ frames are ground truth image latent $\boldsymbol{s}_{g t}^{1: k}$, and the remaining frames $\boldsymbol{s}^{k+1: F}$ are padded with zeros. 
Meanwhile, one-channel masks $\bm{m}^{1:F}$ are used to indicate whether the model should reconstruct the given image latent correspondingly.

\begin{equation}
\begin{aligned}
\boldsymbol{s}^{1: k}=\boldsymbol{s}_{g t}^{1: k},  \quad \boldsymbol{s}^{k+1: F}=0, \quad  \boldsymbol{m}^{1: k}=1, \quad \boldsymbol{m}^{k+1: F}=0, \\
where \quad  k \sim Uniform \{0,1, \ldots,\lfloor F / 4\rfloor\},
\end{aligned}
\end{equation}

This approach differs from using denoised latent codes $\bm{s}^{1:F}_{d}$ during inference, which we find leads to a decrease in video quality. To bridge this domain gap, we introduce a small amount of Gaussian noise to the ground truth image latent during training, simulating the denoised latent:
\begin{equation}
\begin{aligned}
\boldsymbol{s}_{g t, \text { noise }}^{1: k}=\boldsymbol{s}_{g t}^{1: k}+\mathcal{N}\left(0, \sigma^2\right),
\end{aligned}
\end{equation}

\begin{figure}[tp]
    \centering
    \includegraphics[width=1.0\linewidth]{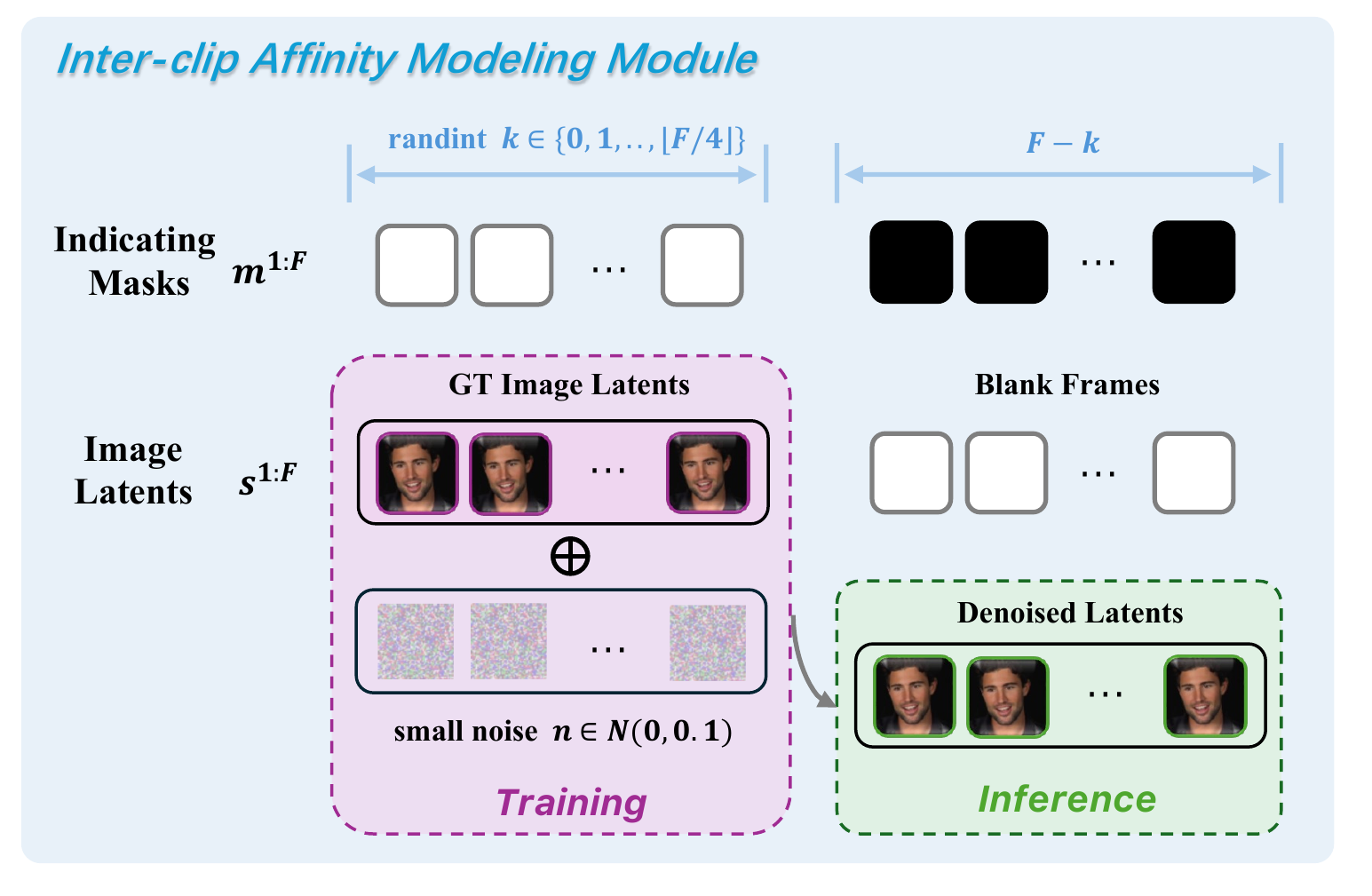}
    \vspace{-3mm}
    \caption{
    \textbf{Illustration of our Inter-clip Affinity Learning Module.}
    The model learns inter-clip temporal consistency by conditioning the image latent of the preceding frames and using masks to indicate whether reconstruction is required.
    }
    \label{fig:interclip}
    \vspace{-1em}
\end{figure}

Here, $\mathcal{N}\left(0, \sigma^2\right)$ represents Gaussian noise with mean 0 and standard deviation $\sigma$ (set to 0.1 here).
Inspired by inter-frame affinity in PIA, we condition the image latent $\bm{s}^{1:F}$ in the input of the U-Net using a lightweight learnable single-layer convolution $\boldsymbol{W}_{cond}$ as the condition module, without compromising the original functionality, as shown in \cref{fig:interclip}.

\begin{figure*}[tp]
    \centering
    \includegraphics[width=0.93\linewidth]{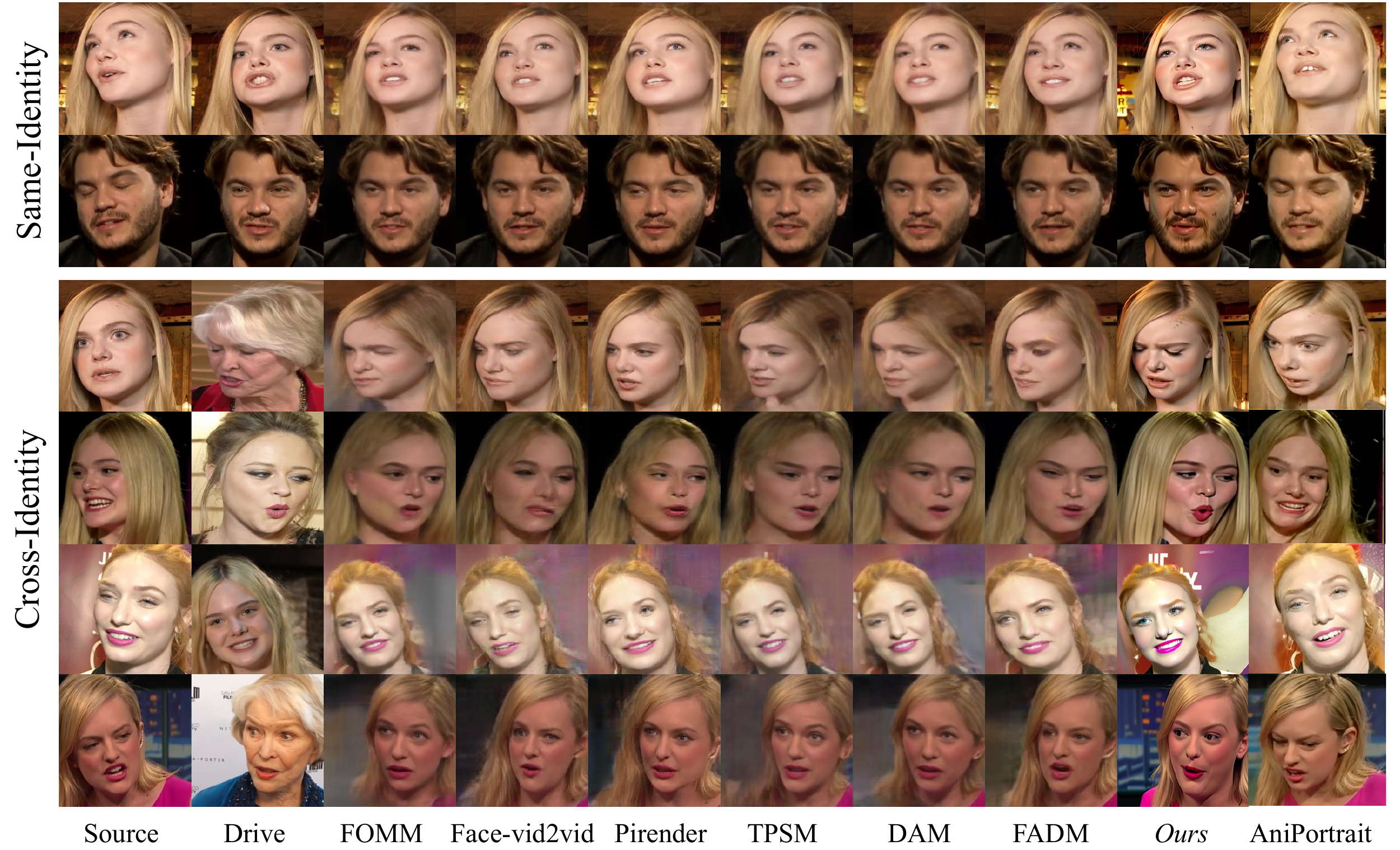}
    \vspace{-1mm}
    \caption{
        \textbf{Same-identity and cross-identity reenactment results on Voxceleb1 test set.}
    } 
    \label{fig:vox}
\end{figure*}

\section{Experiment}
\label{sec:experiment}



%
\noindent
\textbf{Datasets.}
We train our model on VoxCeleb2~\cite{voxceleb1}.
%
We evaluate the VoxCeleb1 test set following the sampling strategy of the PIRenderer~\cite{pirenderer}.
We test on FFHQ~\cite{stylegan} to verify the generalization ability.

\noindent
\textbf{Metrics.}
We use PSNR and LPIPS~\cite{lpips} to evaluate the reconstruction quality for same-identity reenactment. 
Expression, pose, and gaze accuracy are assessed by calculating the average Euclidean distance of the corresponding embedding between the generated and driving faces. These three embeddings are derived through the respective estimator.
%
Identity preservation cosine similarity (CSIM) is calculated by~\cite{curricularface}.
FID is used to evaluate the realism of the generated faces.

\noindent
\textbf{Training Details.}
Our training process is divided into two stages. In the first stage, we train the image-driven model. We begin training from the StableDiffusion v1-5 model and OpenAI clip-vit-large-patch14 vision model. Our models are trained for 30k steps on 4 NVIDIA A100 GPUs, with a constant learning rate of 1e-5 and a batch size 32. To facilitate classifier-free guidance sampling, we train the model without appearance conditions on 10 of the instances. In the second stage, we train the video-driven model. We freeze the SD U-Net and the appearance encoder and train the condition module $\boldsymbol{W}_{cond}$, along with the motion module. Our models are trained for 30k steps on 4 NVIDIA A100 GPUs, with a constant learning rate of 1e-5, a batch size of 8, and a clip sequence of 12.

\begin{table*}[h]
\caption{\textbf{Quantitative evaluations among current popular methods on Voxceleb1 test set.}}
\resizebox{\textwidth}{!}{%
\begin{tabular}{@{}lcccccccccccc@{}}
\toprule
\multirow{2}{*}{Methods} & \multicolumn{7}{c}{Same-Identity}                           & \multicolumn{5}{c}{Cross-Identity}      \\ \cmidrule(l){2-8}\cmidrule(l){9-13} 
                         & PSNR$\uparrow$ &LPIPS$\downarrow$& Exp$\downarrow$&Pose$\downarrow$&Gaze$\downarrow$&CSIM$ \uparrow$&FID$\downarrow$&Exp$\downarrow$&Pose$\downarrow$&Gaze$\downarrow$&CSIM$\uparrow$& FID$\downarrow$\\ \midrule
FOMM~\cite{fomm}         & 22.38          & \underline{0.1405}     & 2.77          & 0.0261         & 0.0554         & 0.8328        & 26.69         & 6.28          & 0.0638         & 0.0959                        & 0.5642          & 42.76         \\
PIRenderer~\cite{pirenderer}     & 21.00          & 0.1468          & 2.94          & 0.0496         & 0.0800         & 0.7997        & \underline{25.48}& \underline{5.84}& 0.0752         & 0.0977                        & 0.5659       & \textbf{35.99}         \\
Face-vid2vid~\cite{face-vid2vid}& 22.63          & \textbf{0.1243}  & 2.84          & 0.0283         & 0.0850         & \underline{0.8383}  & \textbf{25.36}  & 6.68          & 0.0847         & 0.1220                        & 0.6328          & \underline{39.87}         \\
TPSM~\cite{tpsm}     & \underline{23.24}   & 0.1442          & \underline{2.58} & \underline{0.0224}       & \underline{0.0538} & 0.8277        & 33.63         & 6.10          & \underline{0.0535}    & \underline{0.0900}                        & 0.5836          & 50.43         \\
DAM~\cite{dam}        & \textbf{23.37}  & 0.1550          & 2.81          & 0.0263         & 0.0628         & 0.8333        & 36.40         & 6.31          & 0.0626         & 0.0967                        & 0.5534          & 54.13         \\
FADM~\cite{fadm}       & 22.36          & 0.1425          & 2.95          & 0.0303         & 0.0879         & 0.8352        & 31.70         & 6.71          & 0.0821         & 0.1242                        & \underline{0.6522}          & 42.22         \\
Ours                     & 18.64          & 0.1907          & \textbf{2.21} & \textbf{0.0195}& \textbf{0.0482}&\textbf{0.8412}& 60.91         & \textbf{5.03} & \textbf{0.0503}& \textbf{0.0614} & \textbf{0.6778} & 64.49         \\ \bottomrule
\end{tabular}%
}
\label{tab:vox}
\end{table*}



\begin{figure*}[th]
\centering
\includegraphics[width=0.9\linewidth]{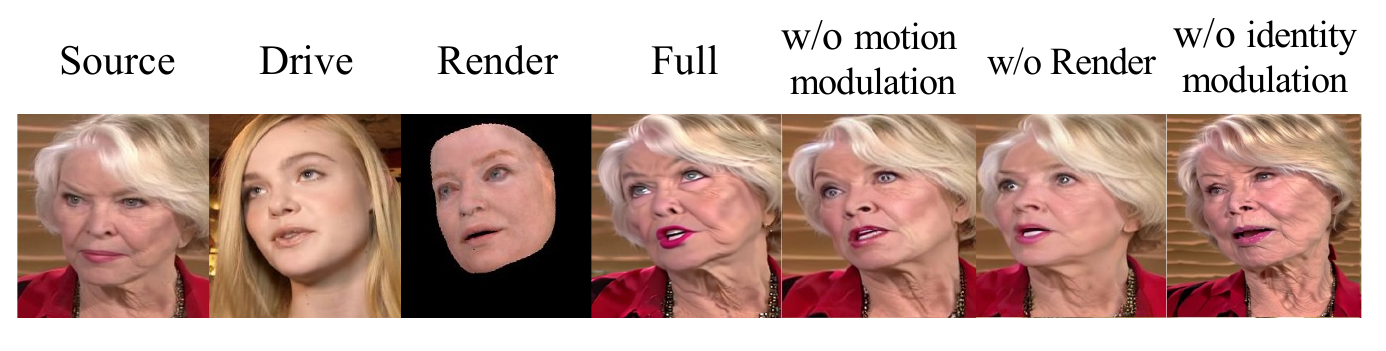}
\vspace{-1.5em}
\caption{
\textbf{Ablation study of different components on the VoxCeleb test set.} 
} 
\label{fig:dr_ablation}
\end{figure*}

\begin{figure}[th]
\centering
\includegraphics[width=1.0\linewidth]{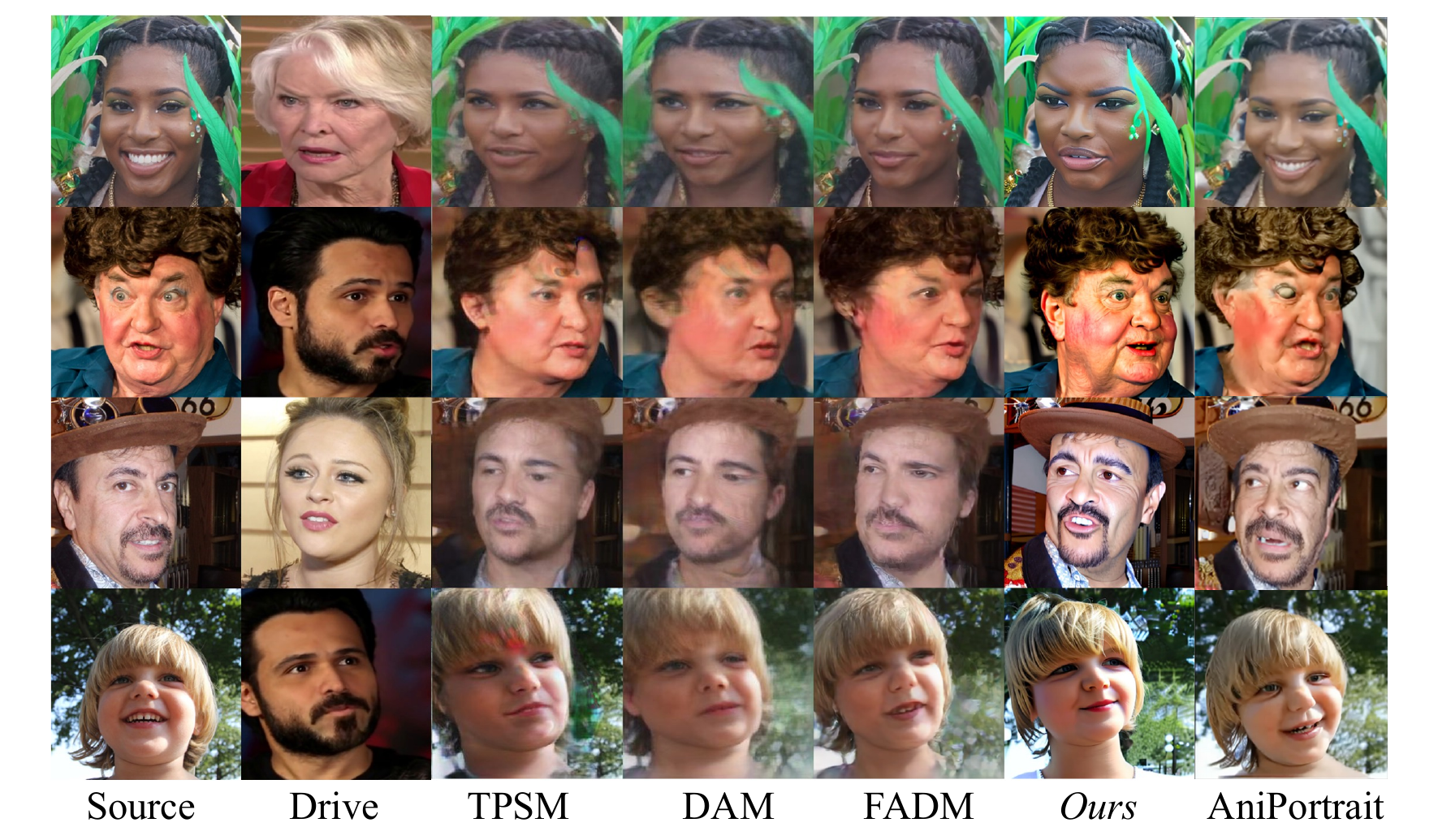}
\caption{
\textbf{Evaluation on FFHQ.} 
Our method exhibits outstanding generalization  capabilities. 
Even when dealing with unseen identities, it maintains consistent identity preservation.
Moreover, it ensures intricate facial textures and precise expressions.
} 
\label{fig:celeba}
\end{figure}


\begin{figure}[h]
\centering
\includegraphics[width=0.99\linewidth]{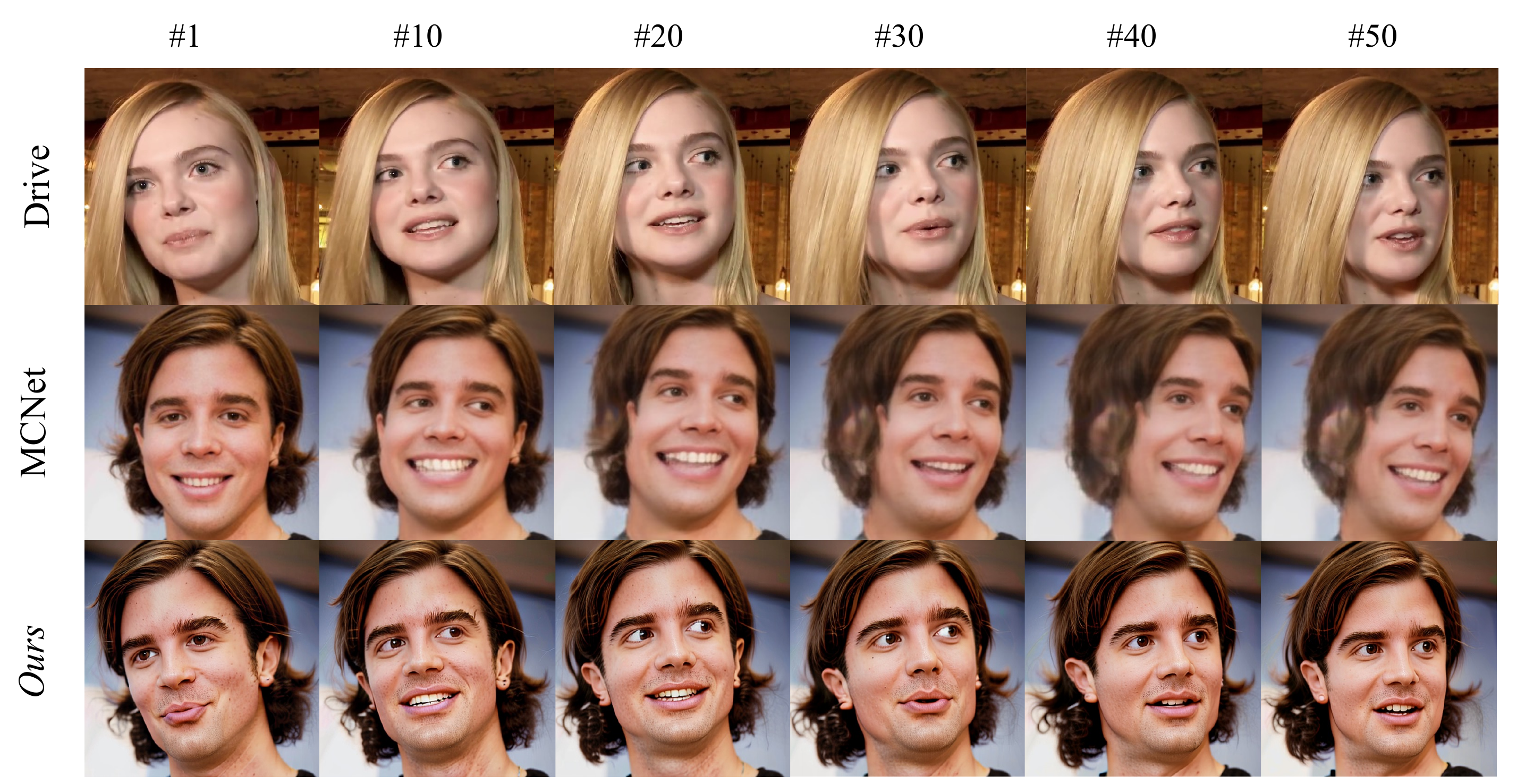}
\caption{
\textbf{Video-driven Reenactment Results}. 
} 
\label{fig:mcnet}
\end{figure}


\subsection{Comparison with State-of-the-Art Methods}

\noindent
\textbf{Methods.}
For image-driven reenactment, we compare our method with GAN-based methods, including FOMM~\cite{fomm}, Face-vid2vid~\cite{face-vid2vid}, PIRenderer~\cite{pirenderer}, TPSM~\cite{tpsm}, DAM~\cite{dam} and diffusion-based methods, including FADM~\cite{fadm} and AniPortrait~\cite{aniportrait}. 
Except AniPortrait, all of these models are trained on VoxCeleb1.
For video-driven reenactment, we compare with MCNet~\cite{mcnet}.
%
%
%

\noindent
\textbf{Qualitative Results.}
In \cref{fig:vox}, previous GAN-based methods tend to produce blurry results with noticeable artifacts. 
While FADM refines the results of FOMM or Face-vid2vid using diffusion, enhancing the generation quality, this approach inherently inherits the motion deviations from these methods.
In contrast, our method dlivers results of both high fidelity and precise control simultaneously. 
Compared to previous outcomes that appear relatively smooth, it is worth noting the nuanced skin wrinkles and light-shadow variations caused by expressions in our method, which make the generated results appear much more lifelike.

\noindent
\textbf{Quantitative Results.}
In \cref{tab:vox}, according to FADM, the data quality of VoxCeleb is relatively subpar, characterized by low resolution and blurred textures. 
Diffusion-based methods tend to generate fine-detailed images, leading to a mismatch between them. 
Consequently, we face noticeable disadvantages in pixel-level metrics (PSNR and LPIPS), and distribution-level metrics (FID). 
However, in terms of semantic-level metrics, including identity similarity (CSIM) and motion accuracy (expression, pose, and gaze), we exhibit clear advantages.


\noindent
\textbf{Handling Unseen Identities.}
In \cref{fig:celeba}, we evaluste on FFHQ to verify the generalization capability of  our approach. 
Whereas previous methods yield blurry results, our approach consistently delivers high-fidelity generation. 



\subsection{Video-driven Reenactment}
MCNet is a method designed specifically for videos and fail when driven by a single image or a short video. 
Thus, we compare our results exclusively with theirs in the context of video generation in \cref{fig:mcnet}. 
MCNet relies on a memory mechanism to maintain clarity in appearance under continuous large-motion changes in videos. However, if the drive image and source image are initially misaligned, the errors in pose and expression will gradually accumulate, leading to a decline in generated quality and the appearance of noticeable artifacts.
In contrast, our approach can produce more accurate motion.


\subsection{Ablation Studies}
We perform ablation studies in \cref{fig:dr_ablation} to verify the effectiveness of the Motion-Identity Modulated Appearance Learning Module (MIA).

\noindent
\textbf{Motion Modulation.}
%
If we remove the render image and rely solely on 3DMM coefficients for motion control, it results in a dramatic decrease in the precision of the generated pose and expression. 
because the model loses a direct reference to the spatial position, making the learning process more challenging.
If we remove the 3DMM coefficients and rely solely on the render image for motion control, it results in a moderate decrease in the precision of the generated pose and expression. 

\noindent
\textbf{Identity Modulation.} Without incorporating the identity contrastive loss, the model, during finetuning, is inclined by the denoise loss to capture low-level features, leading to a decline in identity discrimination capability.

\section{Conclusion}
\label{sec:conclusion}
This work proposes a novel MIMAFace framework to address the limitations of current diffusion-based face animation methods. 
We introduce a Motion-Identity Modulated Appearance Learning Module (MIA) that modulates CLIP features at both motion and identity levels, and an Inter-clip Affinity Learning Module (ICA) to model temporal relationships across clips. 
Our method ensures precise facial motion control and faithful identity preservation, and generates animation videos with intra/inter-clip temporal consistency. 
Extensive experiments demonstrate the effectiveness of the proposed method.

\noindent\textbf{Limitation and Future Works.} Integrating large language models and face-specific SD can enhance performance and application value. 
However, since our approach can generate realistic images that could be used for facial forgery, regulatory constraints are necessary to mitigate this risk.

\clearpage

\section{Appendix}

\subsection{Details of 3DMM}
We choose motion coefficients and the rendered image obtained by 3D Morphable Models (3DMMs) as our unified motion intermediate representation.
Specifically, we employ D3DFR~\cite{d3dfr}, which utilizes ResNet50~\cite{resnet} to predict 3DMM coefficients. These coefficients consist of identity $\boldsymbol{\alpha} \in \mathbb{R}^{80}$, expression $\boldsymbol{\beta} \in \mathbb{R}^{64}$, texture $\boldsymbol{\delta} \in \mathbb{R}^{80}$, illumination $\gamma \in \mathbb{R}^{27}$, and pose $\boldsymbol{\rho} \in \mathbb{R}^6$. Therefore, given an input face $I$, we obtain the coefficient-based face descriptor $P \in \mathbb{R}^{257}$ :
$$
\boldsymbol{P}=\mathcal{\psi}^{I}(\boldsymbol{I})=\{\boldsymbol{\alpha}, \boldsymbol{\beta}, \boldsymbol{\delta}, \boldsymbol{\gamma}, \boldsymbol{\rho}\}.
$$
Given $P$, we acquire the reconstructed 3D face. By projecting it onto the 2D image plane using a fixed renderer $\boldsymbol{R}$, we obtain the image-based face descriptor $\boldsymbol{I}_{R}$:
$$
\boldsymbol{I}_{R}=\boldsymbol{\mathcal { R }}(\boldsymbol{P}) .
$$

\subsection{Temporal Consistency}
Following previous methods ~\cite{mcnet,dagan,fadm}, we primarily focus on comparing the image quality and conducting an ablation study on the MIA module. To highlight the significance of the ICA module in generating smooth videos, we offer supplementary videos that qualitatively demonstrate the temporal stability of our proposed method alongside a quantitative evaluation presented in ~\cref{tab:fvd}. Frechet-Video Distance (FVD)~\cite{fvd1,fvd2} is used to measure the temporal consistency of the generated videos.
GAN-based methods are capable of producing smooth video results without explicit temporal modeling; however, due to their limited generative capacity, they may encounter challenges such as background blurring and facial distortions under large pose variations. In contrast, diffusion-based methods demonstrate excellent generalization and robustness, but their temporal stability is comparatively suboptimal. To maintain temporal stability, these methods often compromise a certain degree of motion control precision.

\begin{table}[h]
\resizebox{\columnwidth}{!}{%
\begin{tabular}{@{}lll@{}}
\toprule
Methods                    & Architecture    & FVD \\ \midrule
MCNet {[}ICCV'23{]} \cite{mcnet}       & GAN       &  616   \\
HyperReenact {[}ICCV'23{]}\cite{bounareli2023hyperreenact} & GAN       & 491 \\
FADM {[}CVPR'23{]} \cite{fadm}        & Diffusion &  285   \\
AniPortrait {[}Arxiv'24{]} \cite{aniportrait} & Diffusion & 341    \\ \midrule
Ours w/o ICA               & Diffusion & 312    \\
Ours                       & Diffusion & 264    \\ \bottomrule
\end{tabular}
}
\caption{Quantitative Evaluation on Temporal Consistency}
\label{tab:fvd}
\end{table}
In our approach, the MIA module delivers robust appearance features, ensuring temporal stability within clips, while the ICA module effectively guarantees smooth transitions between clips. Most importantly, our method also ensures precise motion control, \ie, pose, expression, and gaze.

\subsection{More Results}
\begin{figure}[h]
\centering
\includegraphics[width=1.0\linewidth]{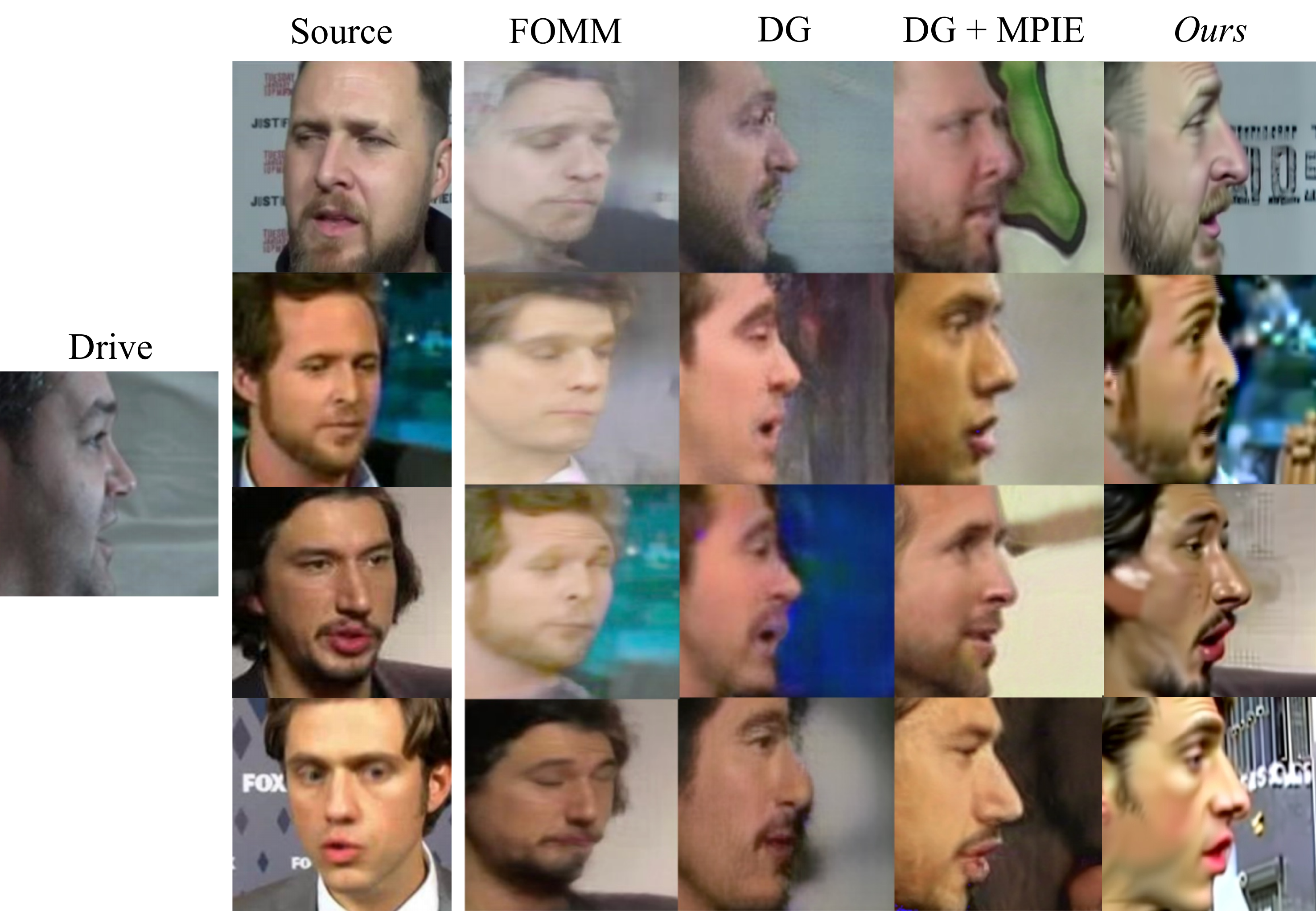}
\caption{
\textbf{Evaluation on MPIE}. 
We compare our method with DG using images directly cropped from the paper. "DG" represents results obtained by training DG on VoxCeleb, while "DG+MPIE" denotes results after fine-tuning DG on MPIE. 
} 
\label{fig:DG}
\end{figure}

\textbf{Handling Extreme Poses}
In \cref{fig:DG}, we evaluate our approach on MPIE~\cite{mpie} to verify the capability to handle extreme poses.
DG~\cite{dg} is a method specifically designed to address large pose reenactment. 
Even without fine-tuning on MPIE, our method achieves better results than DG. 
This is because DG is constrained by the distribution of MPIE, and tends to alter the facial position within images and produces blurry results.

\begin{figure*}[th]
\centering
\includegraphics[width=0.9\linewidth]{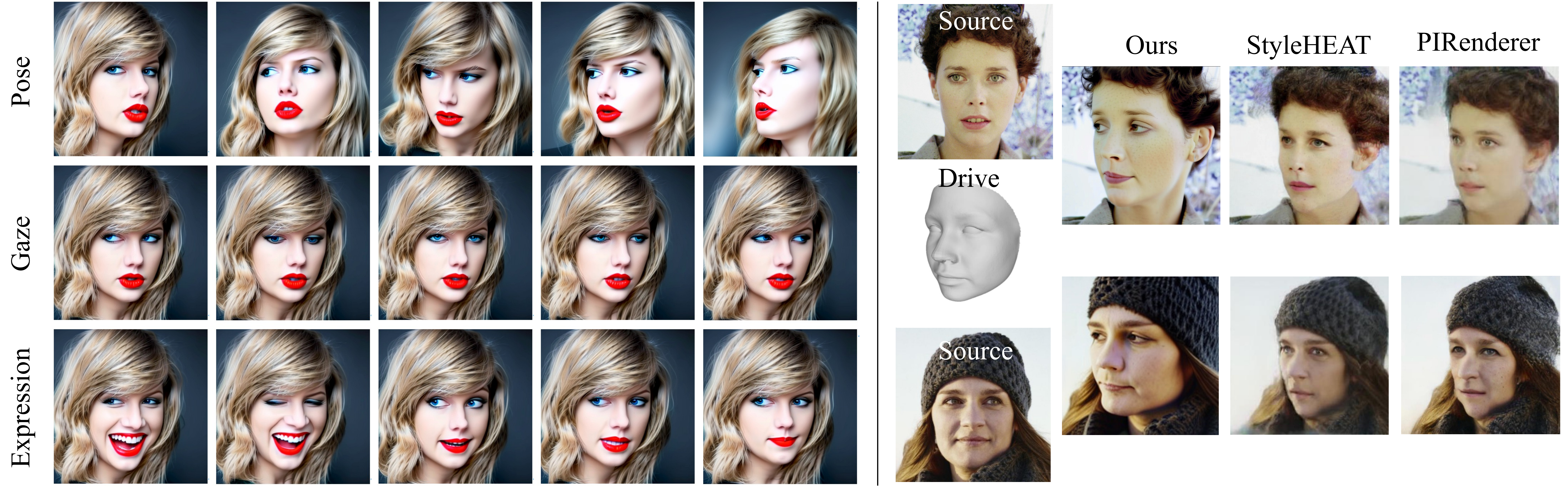}
\caption{
\textbf{Visualization of intuitive facial editing.} 
\textbf{Left:} Results in manipulating pose, expression, and gaze are shown. 
While 3DMM-based methods like PIRenderer and StyleHEAT inherently support pose and expression manipulation, our approach further offers free control of gaze. It exhibits an enhanced capability to decouple and control various motions and is adept at generating intricate appearance details like teeth and hair. 
\textbf{Right:} In comparison to prior methods, our method demonstrates superior structural preservation and clarity, especially under the same pose control, highlighting a marked elevation in the fidelity of facial expression nuances.
}
\label{fig:freecontrol}
\end{figure*}

\begin{figure*}[h]
\centering
\includegraphics[width=0.75\linewidth]{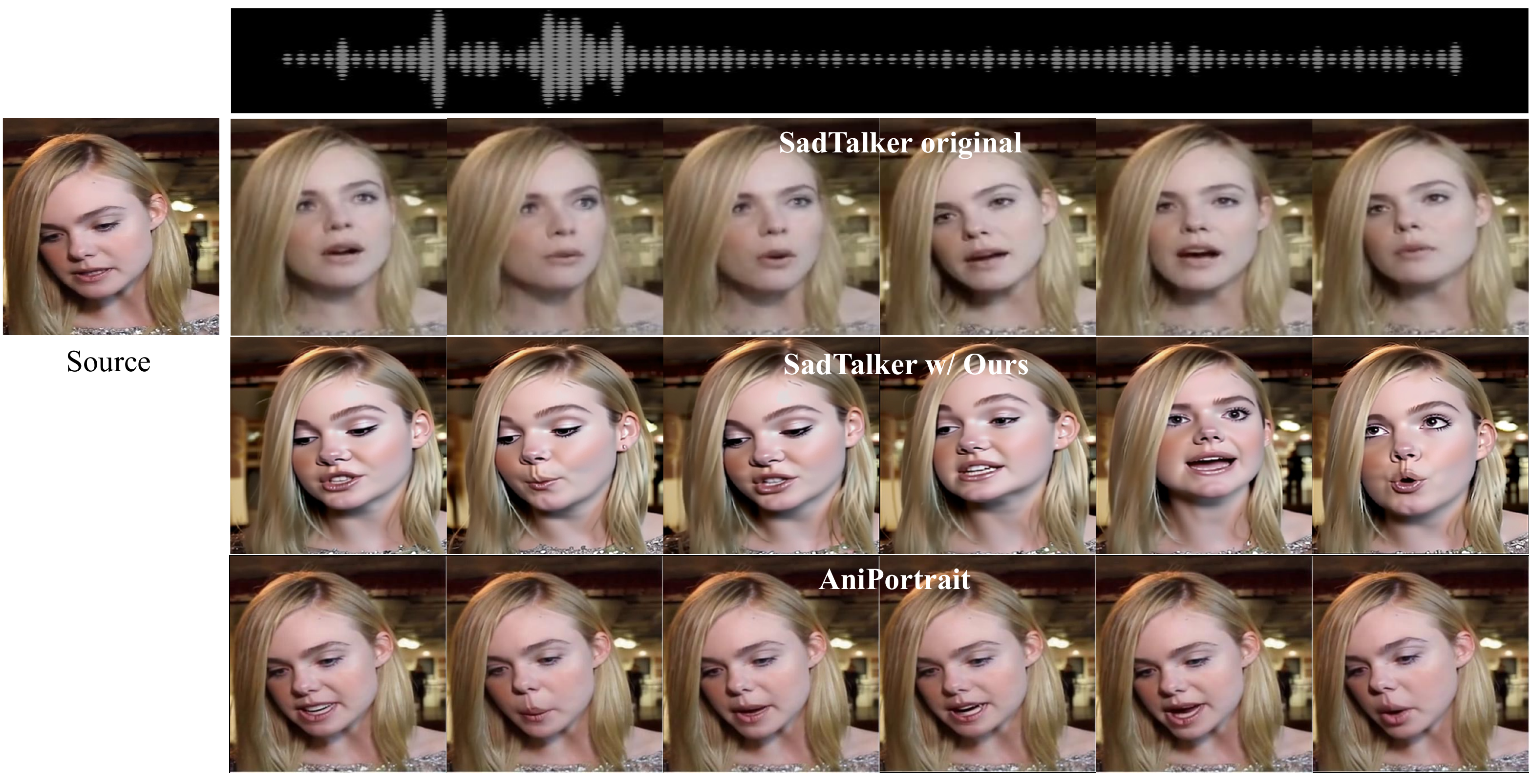}
\caption{
\textbf{Audio-driven Face Animation.}
 Our model is integrated with the audio2exp and audio2pose modules of SadTalker.
} 
\label{fig:withsad}
\end{figure*}

\noindent
\textbf{Audio-driven Face Animation}
Our model can be directly integrated with the audio2exp and audio2pose modules of SadTalker. 
In \cref{fig:withsad}, given that our approach excels in generating fine-grained appearance details under large poses and offers added control over gaze, it yields results that are more vivid and clearer than those produced by the face render in SadTalker and AniPortrait. This demonstrates the advantage of our motion-modulated appearance features in generating subtle facial expressions.

\noindent
\textbf{Intuitive Facial Editing}
3DMM-based methods inherently support free manipulation of pose and expression. Notable techniques include PIRenderer and StyleHEAT. 
Our method also enables the free control of gaze.
\cref{fig:freecontrol}-left displays our results in controlling pose, expression, and gaze. 
Our method decouples and controls various motions to a greater extent and can generate fine-grained appearance details, such as teeth and hair. 
\cref{fig:freecontrol}-right Compared to previous methods, our approach maintains structure and produces clearer images under similar levels of pose control, with a notable improvement in the facial expression details.

\begin{figure*}[h!]
	\centering
	\includegraphics[width=0.99\linewidth]{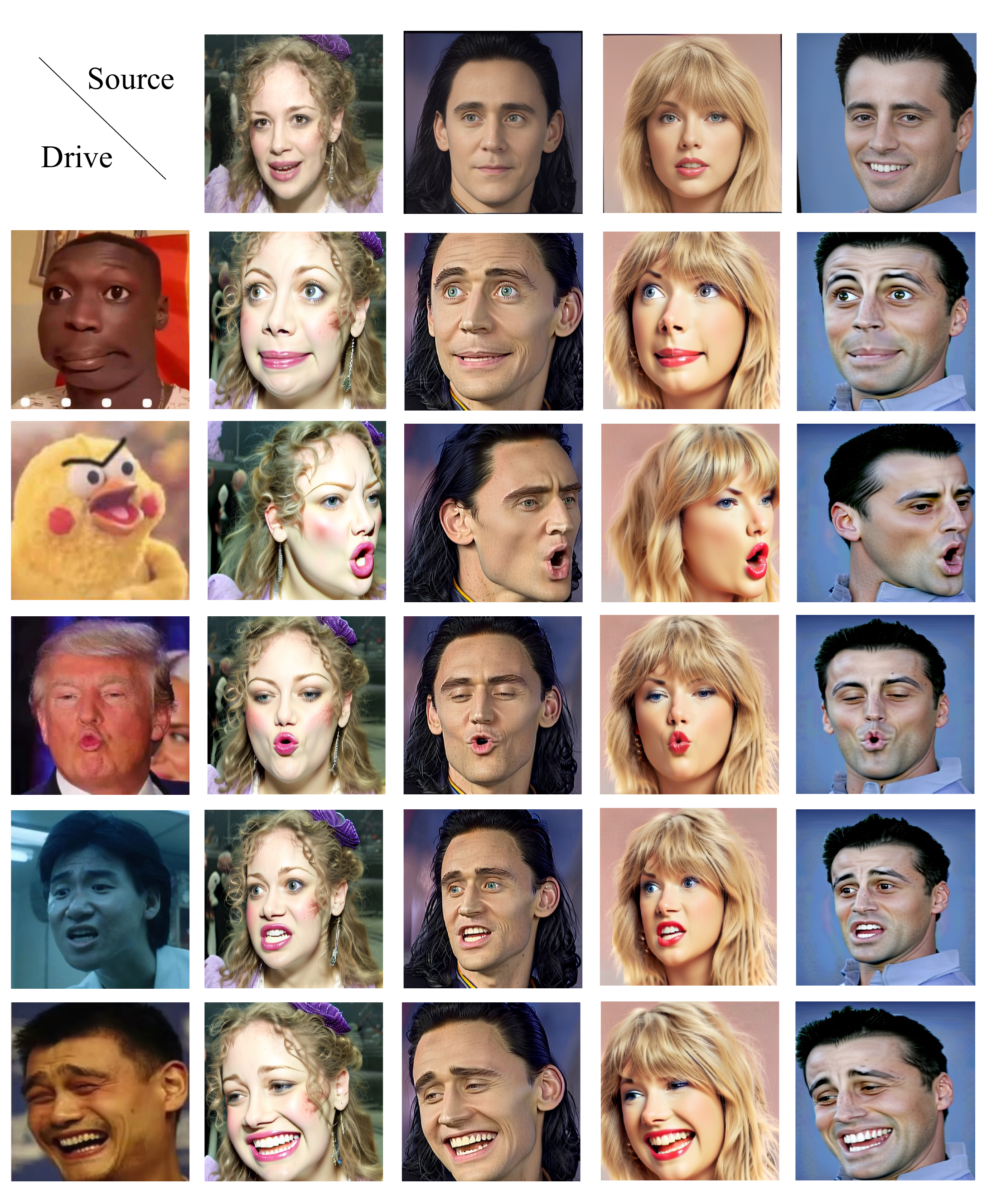}
	\caption{Intriguing Real-World Applications.}
	\label{fig:meme}
\end{figure*}

\noindent
\textbf{Intriguing Real-World Applications}
In \cref{fig:meme}, we showcase intriguing real-world applications. Thanks to the robust generalizability, high generative quality, and precise motion control of our method, we can leverage various amusing meme images from real life to animate specific real-world faces, offering strong entertainment value.

\begin{figure*}[h!]
	\centering
	\includegraphics[width=\linewidth]{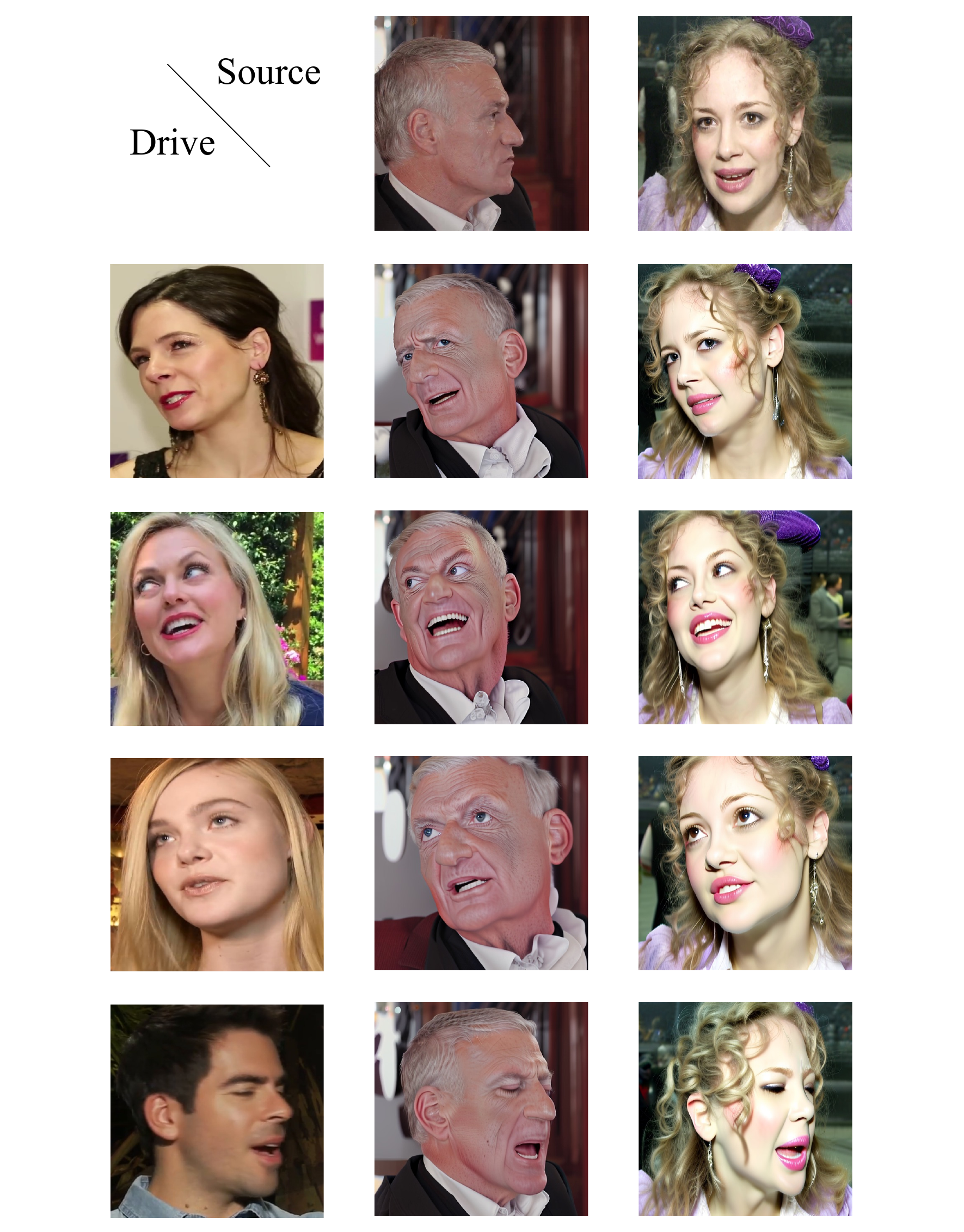}
	\caption{Higher Resolution Results (768*768).}
	\label{fig:vfhq_vis}
\end{figure*}

\noindent
\textbf{Higher Resolution Results (768x768)}
In the main paper, we utilize the VoxCeleb dataset with a resolution of 256 and 512, employing Stable Diffusion 1.5. Here, we showcase visual results obtained from the high-definition VFHQ~\cite{vfhq} dataset at a resolution of 768.
In \cref{fig:vfhq_vis}, even when dealing with a profile source face, our method maintains a rational facial structure, in contrast to prior methods that often yield blurry results. This serves as validation for the efficacy of our approach in handling large-motion reenactment in high-resolution scenarios.

\begin{figure*}[h]
	\centering
	\includegraphics[width=0.99\linewidth]{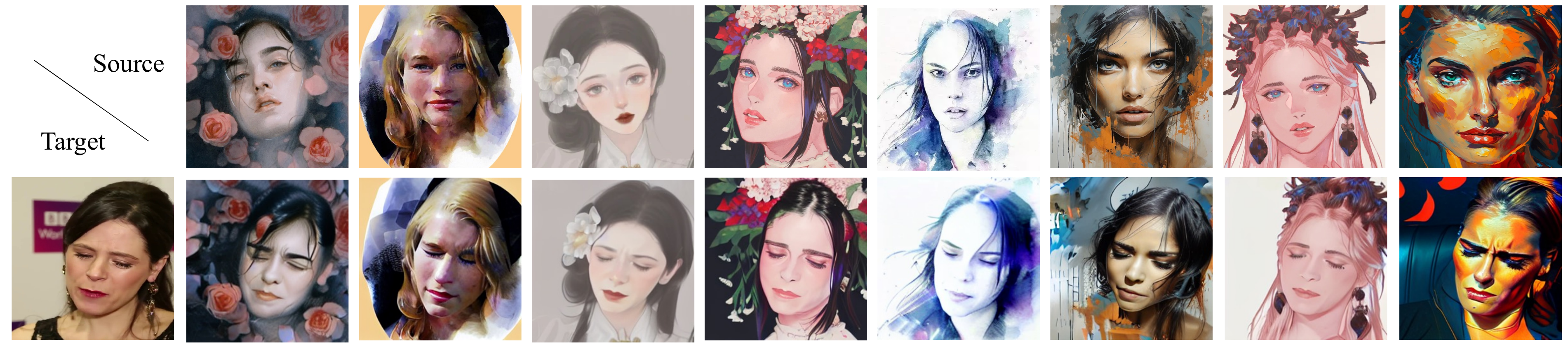}
	\caption{Cross-Domain Reenactment. Real faces effectively reenact artistic facial arts.}
	\label{fig:art}
\end{figure*}

\noindent
\textbf{Cross-Domain Reenactment}
In real-world scenarios, faces in the wild extend beyond authentic photographs of human faces to encompass a diverse array of artistic facial representations. To assess the generalizability of our approach in reenacting source faces across distinct domains, we conduct tests on faces featuring various artistic styles in \cref{fig:art}. 
Our approach adeptly reenacts an array of faces, yielding high-fidelity outcomes. 
This success can be attributed to the pre-trained diffusion model, a capability beyond the reach of previous methods.

{
    \small
    \bibliographystyle{ieeenat_fullname}
    \bibliography{main}
}


\end{document}